\documentclass[10pt, twocolumn, letterpaper]{article}

\usepackage[pagenumbers]{cvpr} % To force page numbers, e.g. for an arXiv version

\usepackage{graphicx}
\usepackage{amsmath}
\usepackage{amssymb}
\usepackage{booktabs}
\usepackage{enumitem}
\usepackage[pagebackref,breaklinks,colorlinks]{hyperref}
\usepackage{svg}

\usepackage[capitalize]{cleveref}
\crefname{section}{Sec.}{Secs.}
\Crefname{section}{Section}{Sections}
\Crefname{table}{Table}{Tables}
\crefname{table}{Tab.}{Tabs.}

\def\cvprPaperID{EECS442} % *** Enter the CVPR Paper ID here
\def\confName{CVPR}
\def\confYear{2024}

\title{DARK: Denoising, Amplification, Restoration Kit}

\author{Zhuoheng Li\\
University of Michigan\\
{\tt\small zhlii@umich.edu}
\and
Yuheng Pan\\
University of Michigan\\
{\tt\small extomato@umich.edu}
\and
Houchen Yu\\
University of Michigan\\
{\tt\small hollinsy@umich.edu}
\and
Zhiheng Zhang\\
University of Michigan\\
{\tt\small alexzh@umich.edu}
}
\begin{document}


\maketitle

%%%%%%%%% ABSTRACT
\begin{abstract}
This paper introduces a novel lightweight computational framework for enhancing images under low-light conditions, utilizing advanced machine learning and convolutional neural networks (CNNs). Traditional enhancement techniques often fail to adequately address issues like noise, color distortion, and detail loss in challenging lighting environments. Our approach leverages insights from the Retinex theory and recent advances in image restoration networks to develop a streamlined model that efficiently processes illumination components and integrates context-sensitive enhancements through optimized convolutional blocks. This results in significantly improved image clarity and color fidelity, while avoiding over-enhancement and unnatural color shifts. Crucially, our model is designed to be lightweight, ensuring low computational demand and suitability for real-time applications on standard consumer hardware. Performance evaluations confirm that our model not only surpasses existing methods in enhancing low-light images but also maintains a minimal computational footprint. The source code is available at \url{https://github.com/hollinsStuart/dark/tree/master}
\end{abstract}

%%%%%%%%% BODY TEXT
\section{Introduction}
\label{sec:intro}
Low-light image enhancement is an important task in computer vision. It is about improving the visibility and quality of images captured in poor light conditions. This enhancement is applied in various fields such as photography, automotive driving systems and medical imaging.

\begin{figure}[ht]
    \centering
    % Row 1
    \begin{minipage}{0.5\linewidth}
        \includegraphics[width=\linewidth]{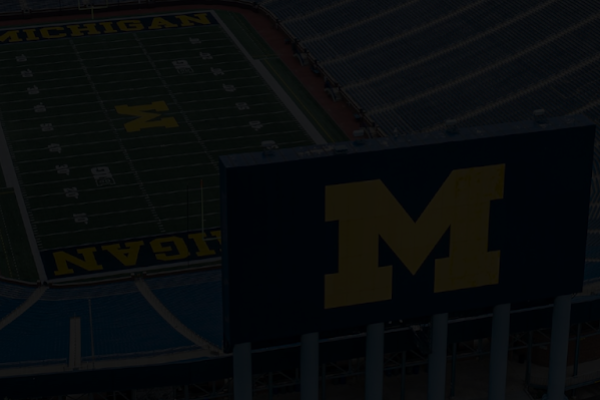}% First image in row 1
    \end{minipage}\hfill
    \begin{minipage}{0.5\linewidth}
        \includegraphics[width=\linewidth]{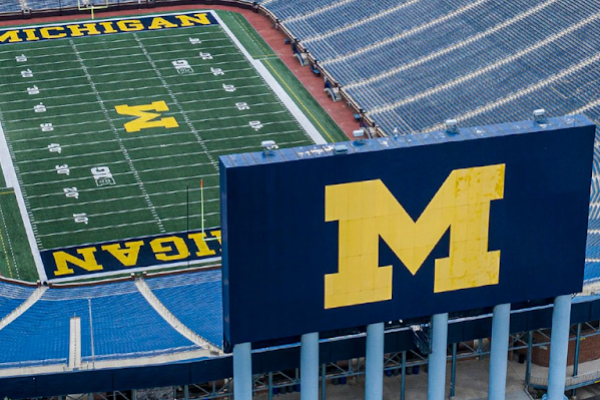} % Second image in row 1
    \end{minipage}\hfill
    \caption{An example of Michigan Stadium}
    \label{fig:um}
\end{figure}

The traditional methods for low-light enhancement, ranging from gamma modification \cite{gamma2016} to histogram equalization \cite{histogram2007} to sophisticated denoising algorithms, often fall short when dealing with the complexity of real-world scenarios. These methods typically operate under constrained assumptions about noise models and lighting conditions, leading to subpar restoration of image details and colors.

In this project, we aim to develop a lightweight architecture for everyday low-light image enhancement. Our model takes advantage of the illumination estimation method \cite{Cai_2023_ICCV} and gets inspired by the MIRNet-v2 \cite{Zamir2022MIRNetv2} model, which can improve visibility while preserving the natural look and details of images, avoiding issues like over-enhancement and unnatural color shifts. It processes images through sequential layers, estimating illumination in each section before enhancing light through parallel blocks that maintain key features.

\section{Background}
\subsection{Plain Methods and Traditional Cognition Methods}
Plain methods such as gamma modification and histogram equalization \cite{gamma2016, histogram2007} often overlook illumination factors, resulting in enhanced images that may perceptually mismatch real-world normal-light scenes. Traditional cognition methods based on Retinex theory \cite{LIME2016guo, wei2018RetinexNet}, although adopting illumination factors and yielding more plausible results, still introduce severe noise and color distortion during enhancement. 

\subsection{Related Work}
% Our model is designed based on MIRNet-v2\cite{Zamir2022MIRNetv2}. This is an image restoration architecture that preserves high-resolution spatial details while incorporating contextual information from multi-scale representations, achieving state-of-the-art results across various image processing tasks. It incorporate parallel multi-resolution convolution streams for feature extraction, facilitate information exchange across resolutions, employ non-local attention mechanisms for contextual information capture and use attention-based multi-scale feature aggregation \cite{Zamir2022MIRNetv2}. However, such a model designed for multiple-purpose results in a relatively huge number of parameters, resource consumption and training time.  MIRNet-v2 has 5.9 million parameters \cite{Zamir2022MIRNetv2}, much more than the 1.88 million parameters in another low-light enhancement work CID-Net \cite{feng2024hvi}. MIRNet-v2 also requires a lot of time for training, 65 hours on SIDD benchmark. Since this paper focuses on low-light enhancement and denoising, our model architecture is a lightweight one that achieves similar results with much fewer resources.
Our model draws inspiration from MIRNet-v2\cite{Zamir2022MIRNetv2}, a cutting-edge image restoration architecture that retains high-resolution details and integrates contextual information from various scales, thereby setting new benchmarks in several image processing tasks. It utilizes parallel multi-resolution convolution streams for extracting features, enables cross-resolution information exchange, and incorporates non-local attention mechanisms and attention-based multi-scale feature aggregation for enhanced contextual understanding. However, MIRNet-v2 \cite{Zamir2022MIRNetv2}, being a multi-purpose model, is resource-intensive, containing 5.9 million parameters and requiring considerable computational resources and training time—specifically, 65 hours on the SIDD benchmark. In contrast, another model designed for low-light enhancement, CID-Net \cite{feng2024hvi}, has significantly fewer parameters at 1.88 million. Given our focus on low-light enhancement and denoising, we have opted for a more streamlined, lightweight model architecture that delivers comparable performance with far less resource usage.

In addition, we implemented an illumination estimator that more effectively estimates the illumination component of an image based on the work done by Retinexformer\cite{Cai_2023_ICCV}. It is a novel one-stage transformer-based method for low-light image enhancement, which revises the traditional Retinex model to effectively handle image corruptions like noise and color distortion. It incorporates an Illumination-Guided Transformer (IGT) to exploit illumination information, improving the modeling of long-range dependencies. A significant improvement in Retinexformer is that it estimates the light-up map with a tensor of 3 channels $\bar{L} \in \mathbb{R}^{H \times W \times 3}$ instead of the traditional single channel approach $L \in \mathbb{R}^{H \times W} $. According to the Retinex theory, a low-light image can be decomposed into a reflectance image $R$ and a light-up mag $L$. 
\begin{equation*}
    I = R \odot L
\end{equation*} 

Then the lit-up image is obtained by element-wise division $(I. / L.)$, where computers are prone to suffer data overflow and minor random errors that results in inaccuracy. Therefore, $\bar{L}$ gives a more robust estimation\cite{Cai_2023_ICCV}. 

\section{Method}

\begin{figure*}[ht]
  \centering
  \includegraphics[width=0.98\linewidth]{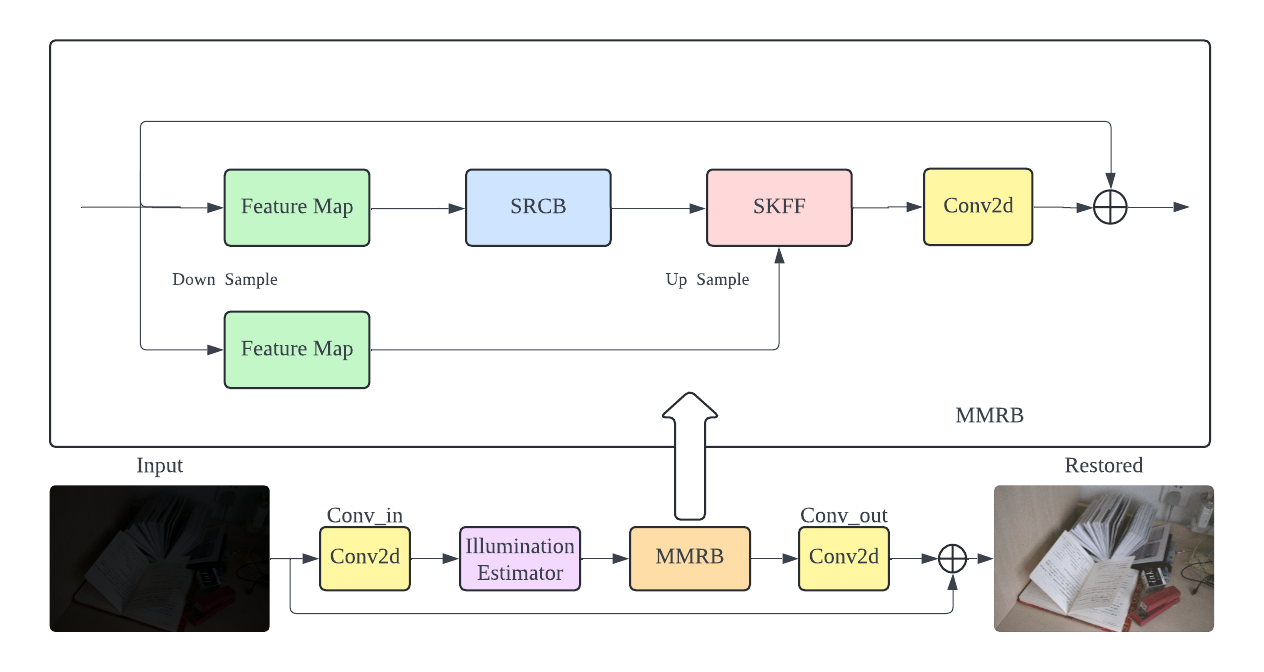}
  \captionsetup{justification=centering}
  \caption{Overview of our proposed method (DARK)\\
  SRCB: Simplified Residual Context Block,
SKFF: Selective Kernel Feature Fusion,
MMRB: Modified Multiscale Residual Block}.
  \label{img:arch}
\end{figure*}

A schematic of the proposed DARK network is shown in Figure \ref{img:arch}. Here are details of the fundamental building blocks of our method, including
the following key elements:
\begin{itemize}
    \item Retinex-based illumination estimator to extract illumination features
    \item Simplified contextual blocks to extract attention-based features
    \item Selective Kernel Feature Fusion to perform aggregation based on self-attention
\end{itemize}

The architecture initializes with a convolutional layer that expands the input image to a predefined feature space. An embedded module, the Illumination Estimator, dynamically adjusts the initial image based on estimated illumination maps. This is followed by a sequence of modules forming the main body of the network, including a Modified Multiscale Residual Block (MMRB) that adapts to varying feature complexities and scales, and additional convolutional layers for deep feature processing.

The network concludes with an output convolution that ensures the integration of enhancements with the original image content. This architecture is optimized for tasks that require detailed control over image illumination and quality, demonstrating versatility and robustness in handling low-light enhancement with minimal parameters.

Note that the network is implemented using PyTorch. Additionally, the OpenCV library is utilized for efficient image processing and manipulation tasks.

\subsection{Retinex-based Illumination Estimator}

The Illumination Estimator, inspired by the Retinex theory and detailed in Retinexformer \cite{Cai_2023_ICCV}, uses convolutional layers to separate an image into its illumination and reflectance components. These layers extract key features from the input image to generate a light-up map and feature, crucial for modeling the complex dynamics of real-world lighting conditions. This approach enhances image clarity under various lighting environments by accurately estimating illumination.

The architecture of the module consists of three principal convolutional layers. The first layer, a $1 \times 1$ convolution (\texttt{conv1}), expands the input features from the number of input channels ($n_{\text{fea\_in}}$) to a higher-dimensional intermediate feature space ($n_{\text{fea\_middle}}$). This transformation is pivotal as it prepares the spatial information for more granular feature extraction without altering the spatial resolution.

Following the initial expansion, a depthwise convolution layer (\texttt{depth\_conv}) with a $5 \times 5$ kernel processes each channel of the feature map independently, maintaining the same number of groups as the intermediate features. This layer's design reduces the model's complexity and parameter count while enhancing the capability to learn nuanced, channel-specific features.

The final transformation in the network is performed by another $1 \times 1$ convolution layer (\texttt{conv2}), which reduces the dimensionality of the feature space from $n_{\text{fea\_middle}}$ to $n_{\text{fea\_out}}$. This layer effectively synthesizes the detailed features processed by the depthwise convolution into a coherent illumination map, reflecting the aggregate lighting conditions of the input image.

In the forward propagation process, the module directly computes a mean channel from the input image by averaging the original channels along the spatial dimensions. This computed channel is then appended to the input image. This augmented input integrates a global illumination context with the local features, enhancing the model's ability to capture comprehensive lighting information. The sequence of transformations through \texttt{conv1}, \texttt{depth\_conv}, and \texttt{conv2} culminates in the production of an illumination map that robustly represents the spatially varying illumination conditions, thereby supporting enhanced image processing applications such as dynamic range compression and color constancy.

\subsection{Simplified Residual Contextual Block}

The Simplified Residual Contextual Block (SRCB) incorporates a streamlined approach to enhancing feature maps with context-sensitive information. This simplification reduces computational overhead while maintaining the effectiveness of the feature enhancement. The SRCB is composed of two primary components: the Simplified Context Block (SCB) and a convolutional body.

\subsubsection{Simplified Context Block (SCB)}

The SCB module is designed to efficiently model the contextual information within the feature maps. Its operation can be summarized as follows:
\begin{itemize}
    \item A convolution layer with a kernel size of $1 \times 1$ computes a context mask directly from the input feature maps. This mask aims to identify regions within the feature map that are more relevant for further processing.
    \item The mask is reshaped and normalized using a softmax function to ensure it represents a probability distribution over the spatial dimensions of the input.
    \item The final context is calculated by applying this mask to the original input through element-wise multiplication followed by a summation across the spatial dimensions. This operation focuses on enhancing features in areas with higher probabilities, effectively incorporating local contextual information into the feature maps.
\end{itemize}

The SCB is characterized by its efficiency and reduction of redundancy, focusing on essential contextual information without extensive computational requirements.

\subsubsection{Simplified Residual Contextual Block (SRCB)}

The SRCB integrates the SCB into a residual learning framework:
\begin{itemize}
    \item The convolutional body of the SRCB consists of a single convolutional layer that processes the input feature maps, maintaining their spatial dimensions.
    \item The processed features are then passed through the SCB, where context-based enhancements are applied.
    \item The output of the SCB is added back to the original input through a residual connection, promoting the flow of gradients during training and preserving the original feature structures while integrating enhanced contextual information.
\end{itemize}

This structure allows the SRCB to efficiently and effectively enhance feature maps by emphasizing informative features dynamically, based on the learned contextual cues, while maintaining computational efficiency.

The SRCB's design reflects an optimal balance between computational efficiency and the capability to enhance feature representation within neural networks, particularly beneficial in scenarios where computational resources are limited.

\subsection{Selective Kernel Feature Fusion}

The SKFF module operates on features from different resolution streams, and
performs aggregation based on self-attention. Typically, feature fusion is achieved through basic methods like concatenation or summation. However, these methods do not fully leverage the network’s capacity, as noted in \cite{Li_2019_CVPR}. Our modified multi-scale residual block (MMRB) introduces a nonlinear method for combining features from various resolution streams by employing a self-attention mechanism, termed selective kernel feature fusion (SKFF), inspired by \cite{Li_2019_CVPR}.

At its core, the SKFF comprises several key components:
An \textbf{Adaptive Average Pooling layer} reduces spatial dimensions to a single value per channel, effectively summarizing the global contextual information of the feature maps.
A \textbf{dimensionality reduction convolution layer} (identified as \texttt{conv\_du} in the implementation) compresses the number of channels from \texttt{in\_channels} to a smaller set determined by the reduction factor, producing a compact feature representation.
A series of \textbf{fully connected layers} (or $1\times1$ convolutions), stored in \texttt{fcs} for each scale. These layers project the reduced feature representation back up to the original number of channels, generating a set of attention maps that signify the importance of each feature map at every scale.
A \textbf{Softmax layer} normalizes these attention maps across the scales, ensuring that the fusion process is adaptively weighted according to the relevance of each scale's features.

During the forward pass, the module first concatenates input features from all scales and reshapes them accordingly. It computes a unified feature map (\texttt{feats\_U}) by summing the scaled inputs, which is then processed through the dimensionality reduction and fully connected layers to produce the scale-specific attention maps. These attention vectors are applied to the original multi-scale feature maps, and the final output feature map (\texttt{feats\_V}) is obtained by summing the attention-weighted features across scales.

This fusion mechanism allows the SKFF module to emphasize more informative features while suppressing less useful ones, adapting to the content of the input features. This capability is particularly beneficial in tasks involving complex visual patterns and varying feature scale importance, such as in high-resolution image processing or detailed scene analysis.

\begin{figure*}[htbp]
    \centering
    % Row 1
    \begin{minipage}{0.245\linewidth}
        \includegraphics[width=\linewidth]{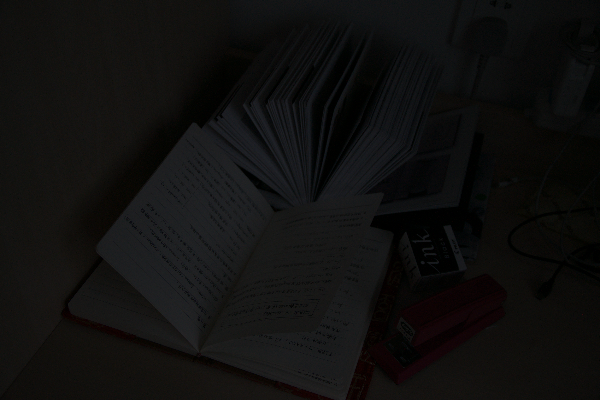}% First image in row 1
    \end{minipage}\hfill
    \begin{minipage}{0.245\linewidth}
        \includegraphics[width=\linewidth]{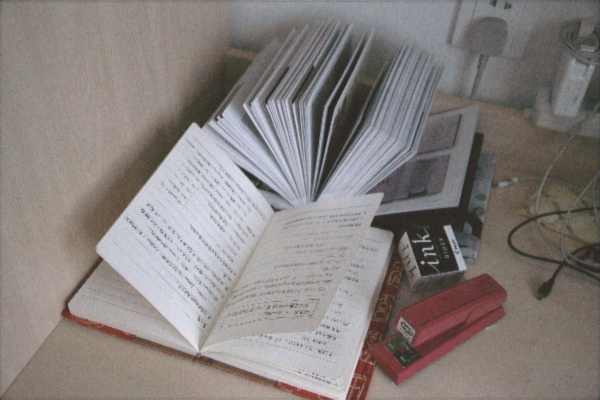} % Second image in row 1
    \end{minipage}\hfill
    \begin{minipage}{0.245\linewidth}
        \includegraphics[width=\linewidth]{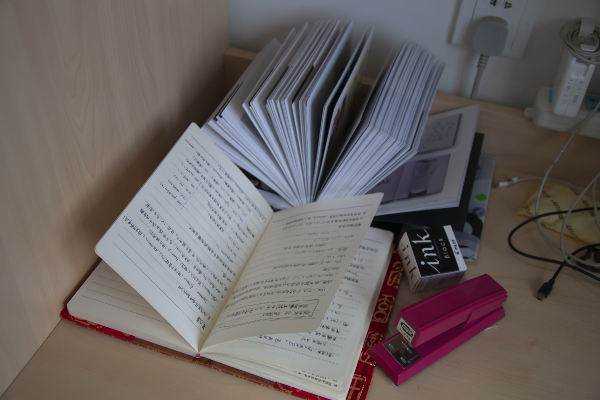} % Third image in row 1
    \end{minipage}\hfill
    \begin{minipage}{0.245\linewidth}
        \includegraphics[width=\linewidth]{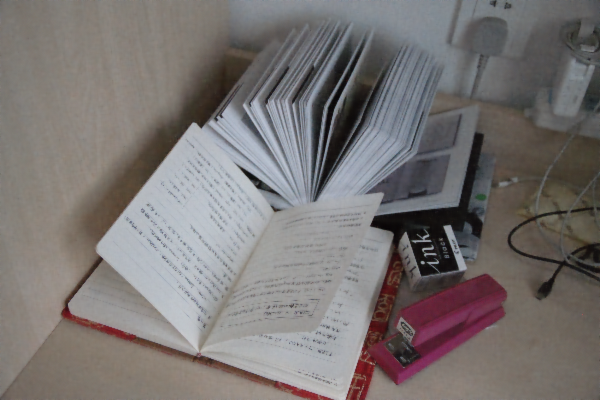} % Third image in row 1
    \end{minipage}

    % Row 2
    \begin{minipage}{0.245\linewidth}
        \includegraphics[width=\linewidth]{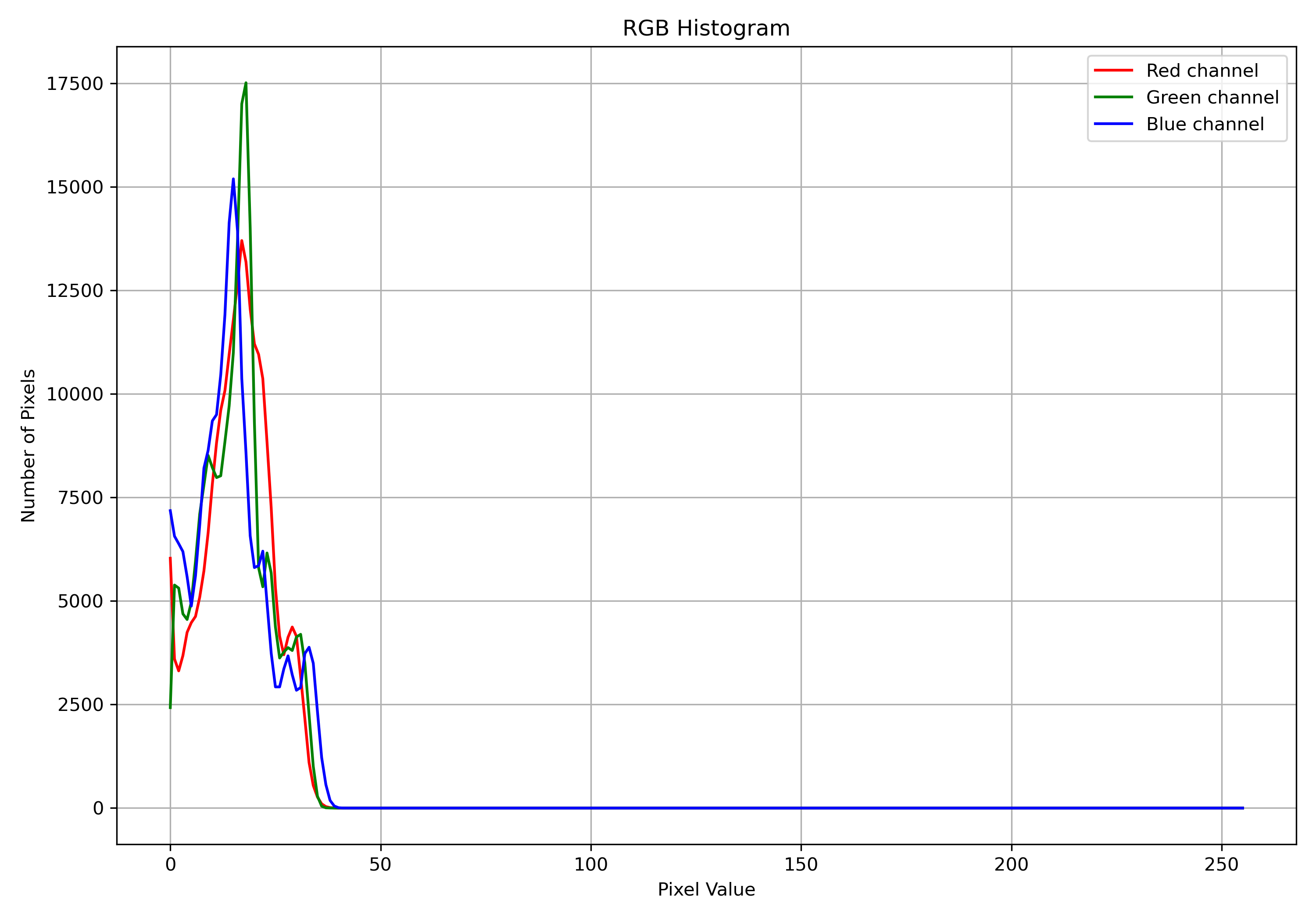} % First image in row 2
    \end{minipage}\hfill
    \begin{minipage}{0.245\linewidth}
        \includegraphics[width=\linewidth]{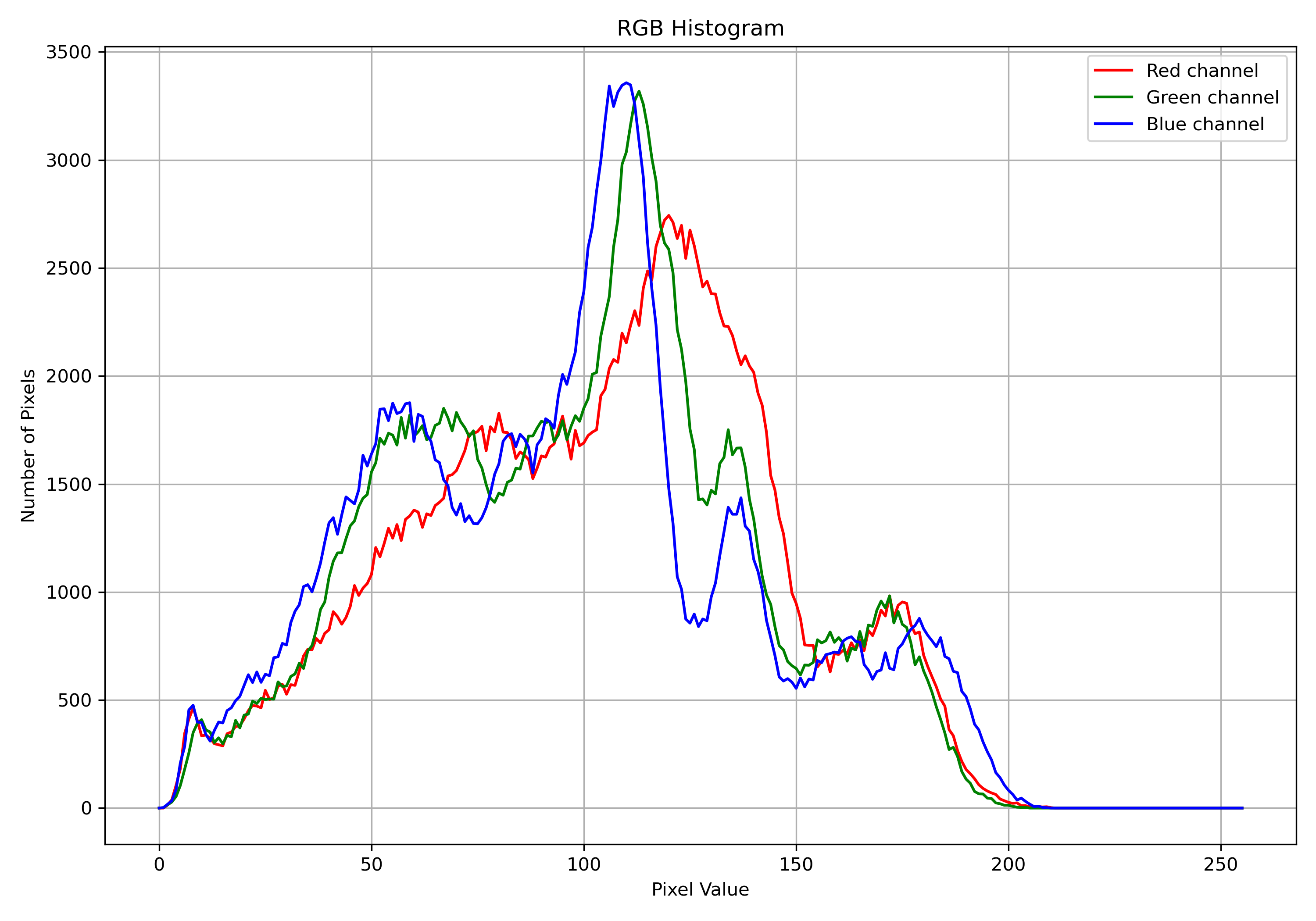} % Second image in row 2
    \end{minipage}\hfill
    \begin{minipage}{0.245\linewidth}
        \includegraphics[width=\linewidth]{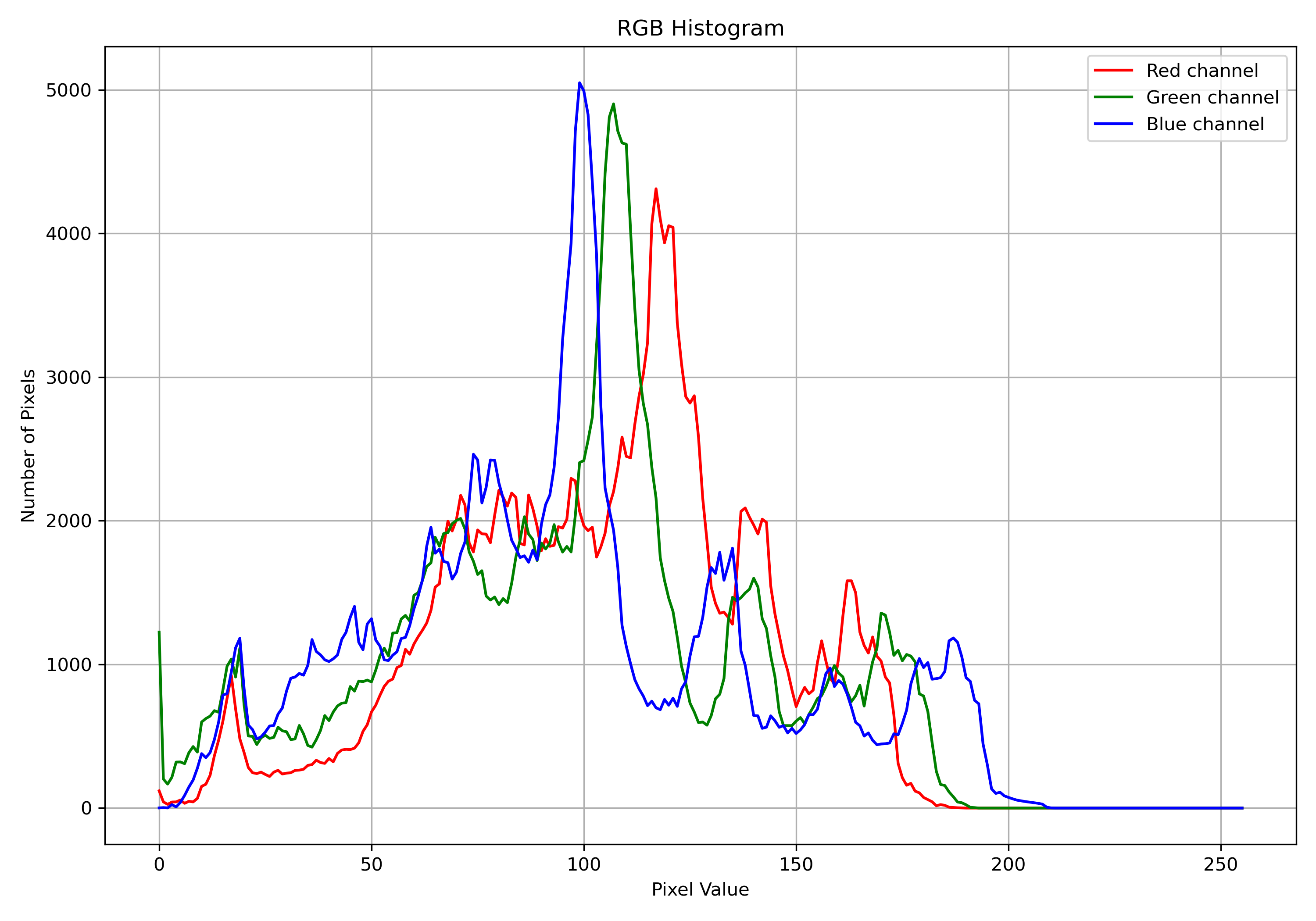} % Third image in row 2
    \end{minipage}\hfill
    \begin{minipage}{0.245\linewidth}
        \includegraphics[width=\linewidth]{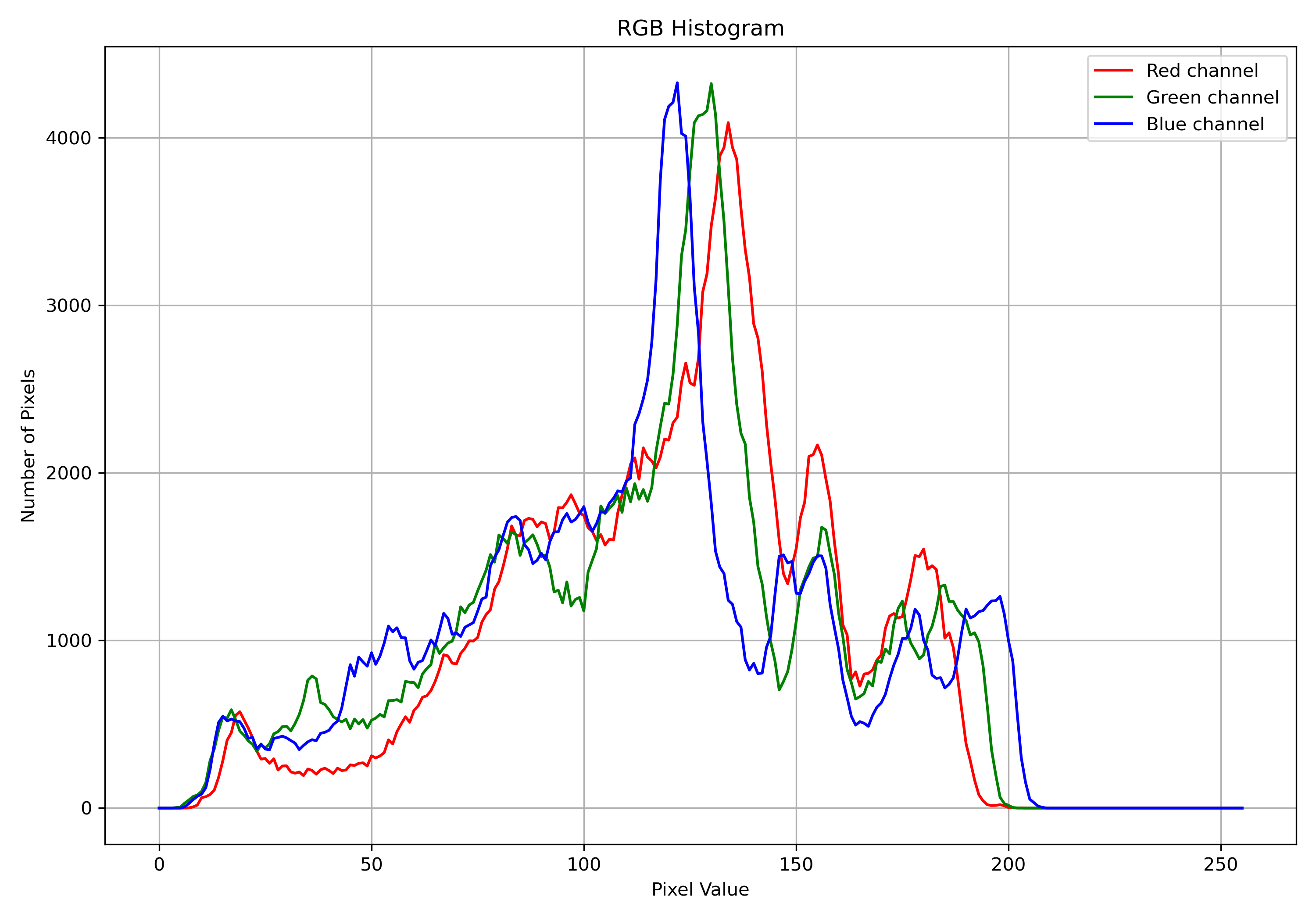} % Third image in row 1
    \end{minipage}

    % Row 3
    \begin{minipage}{0.245\linewidth}
        \includegraphics[width=\linewidth]{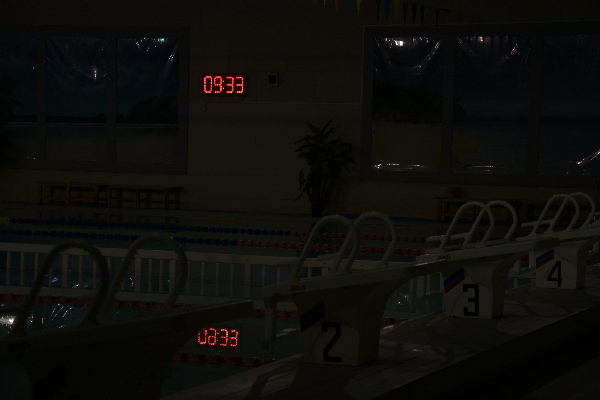} % First image in row 2
    \end{minipage}\hfill
    \begin{minipage}{0.245\linewidth}
        \includegraphics[width=\linewidth]{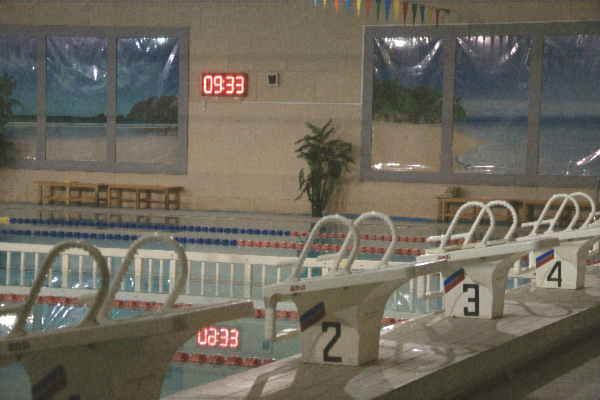} % Second image in row 2
    \end{minipage}\hfill
    \begin{minipage}{0.245\linewidth}
        \includegraphics[width=\linewidth]{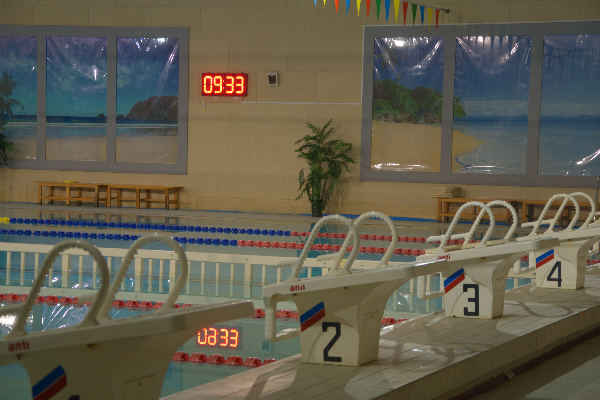} % Third image in row 2
    \end{minipage}\hfill
    \begin{minipage}{0.245\linewidth}
        \includegraphics[width=\linewidth]{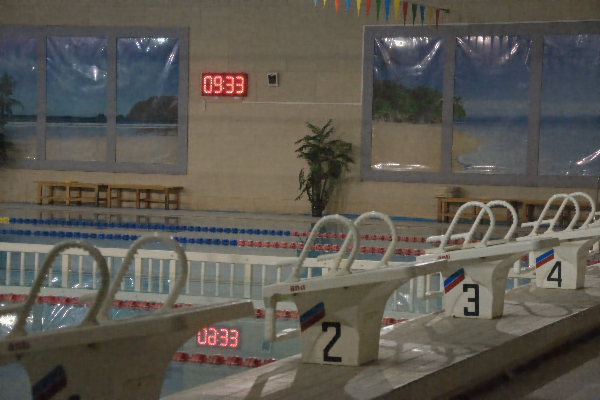} % Third image in row 1
    \end{minipage}
    
    % Row 4
    \begin{minipage}{0.245\linewidth}
        \includegraphics[width=\linewidth]{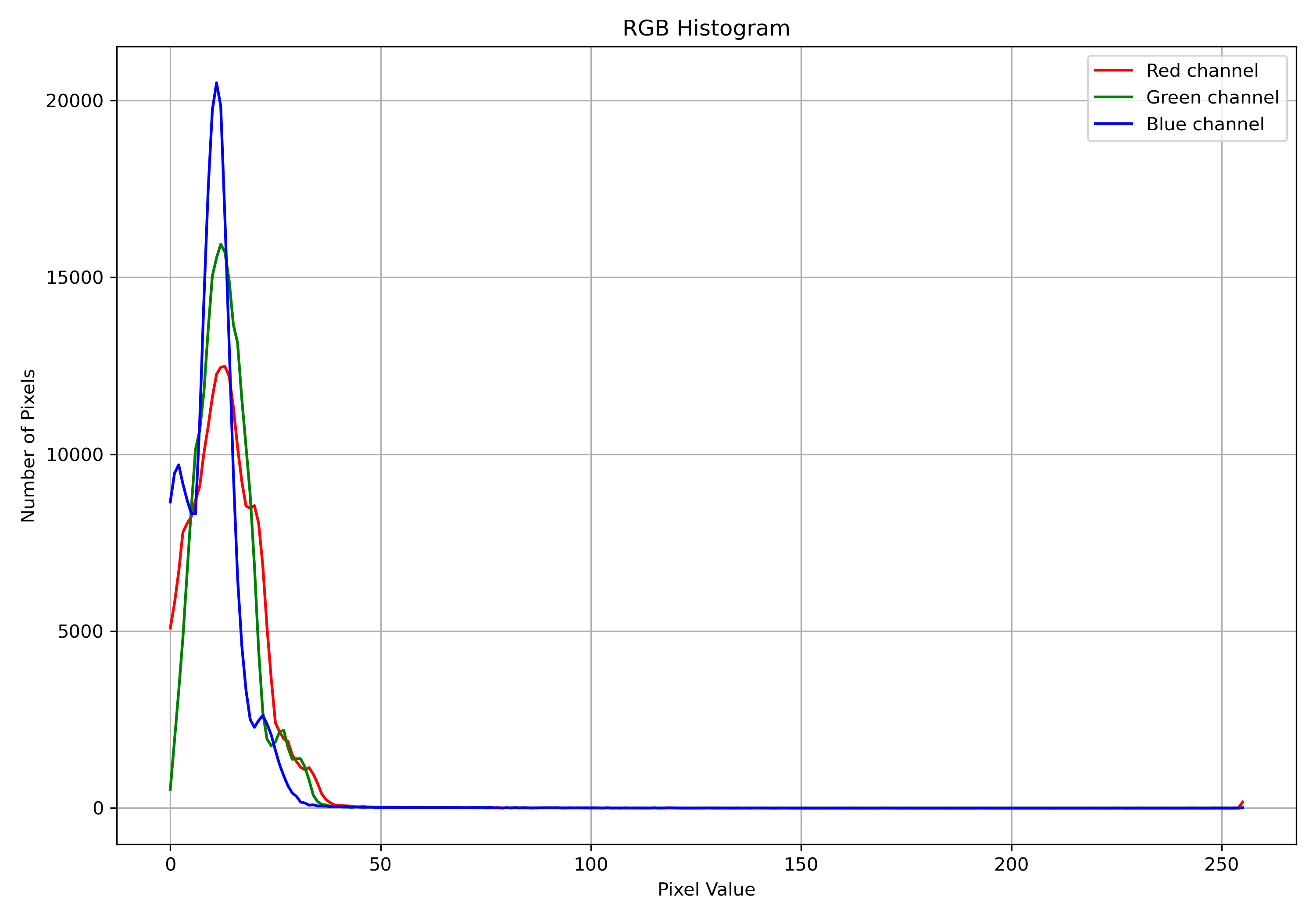} % First image in row 2
        \caption*{Input}
    \end{minipage}\hfill
    \begin{minipage}{0.245\linewidth}
        \includegraphics[width=\linewidth]{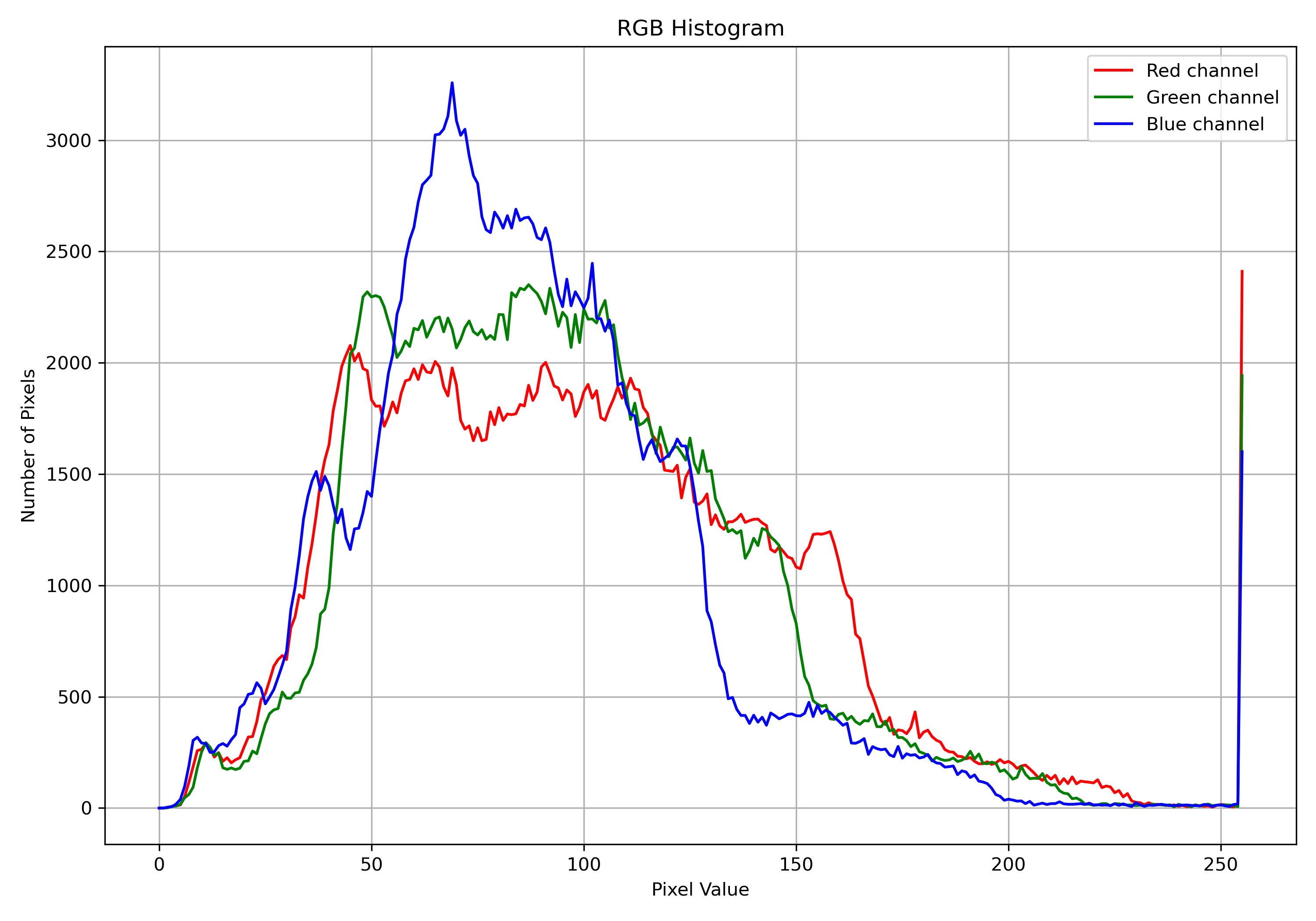} % Second image in row 2
        \caption*{Restored}
    \end{minipage}\hfill
    \begin{minipage}{0.245\linewidth}
        \includegraphics[width=\linewidth]{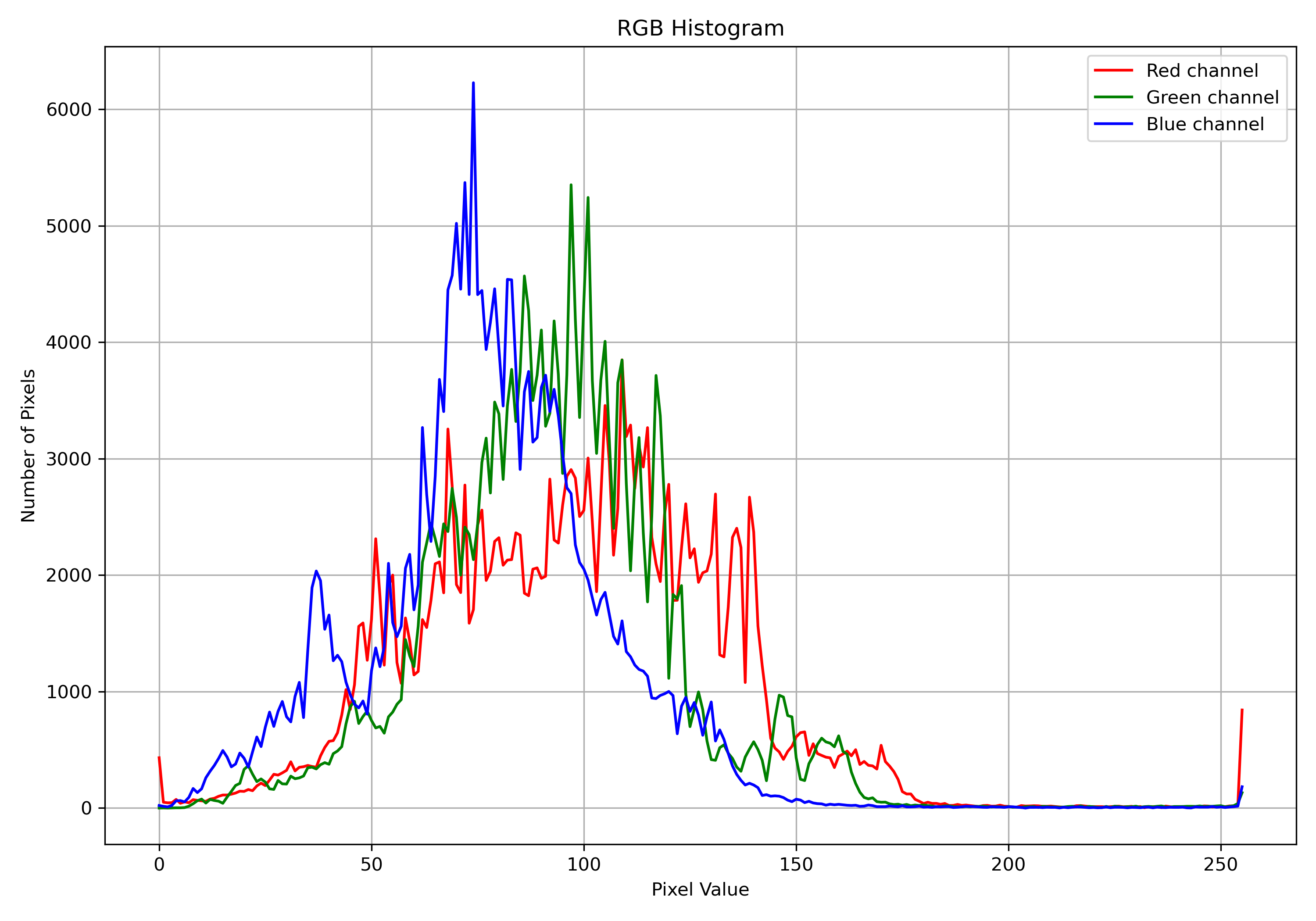} % Third image in row 2
        \caption*{Ground Truth}
    \end{minipage}\hfill
    \begin{minipage}{0.245\linewidth}
        \includegraphics[width=\linewidth]{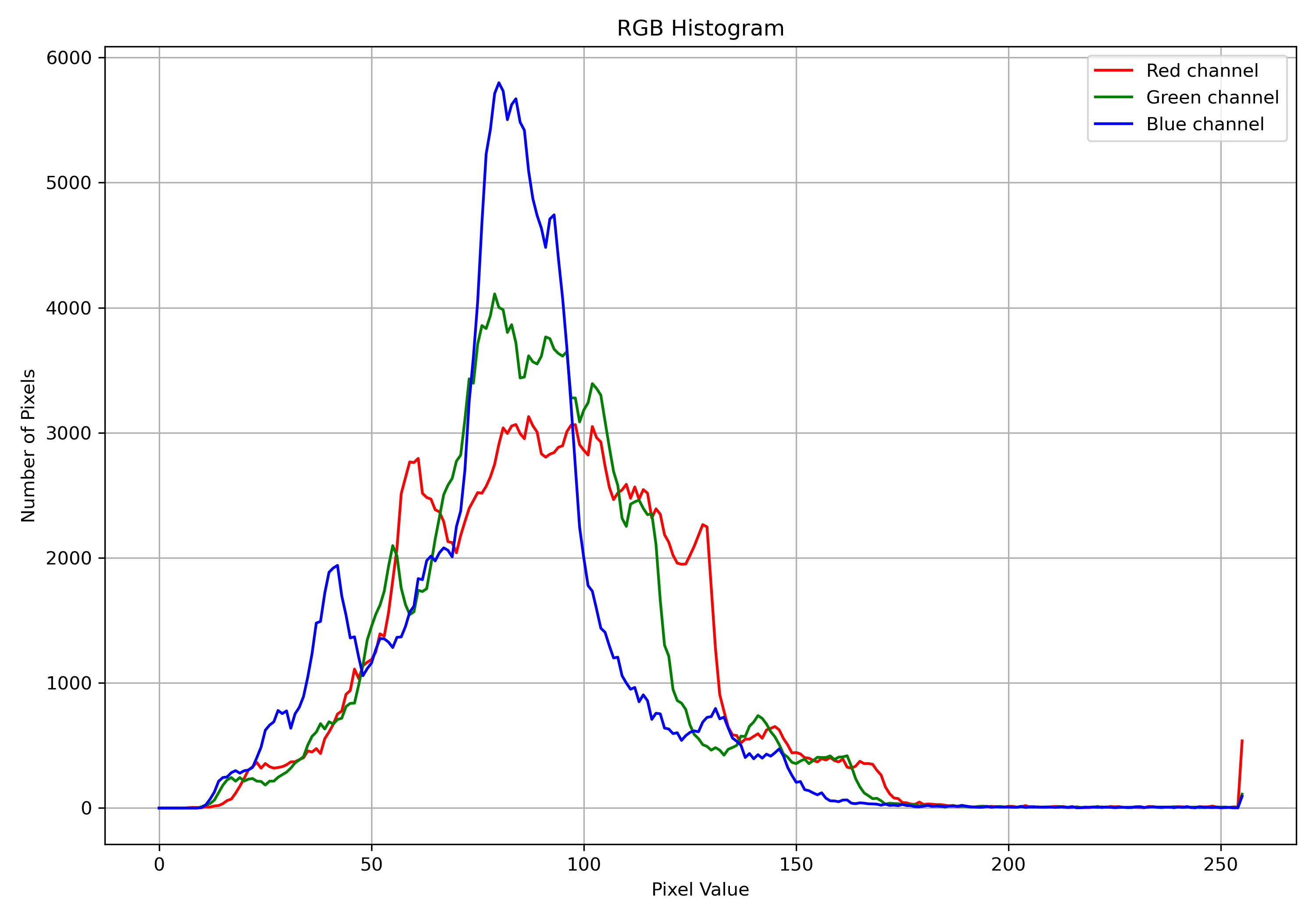} % Third image in row 1
        \caption*{Output from Retinex Former\cite{Cai_2023_ICCV}}
    \end{minipage}
    \caption{Visual results of DARK model on LoL dataset\cite{loldata}. For a better understanding we added the histogram for each image.}
    \label{fig:results_w_plots}
\end{figure*}

\begin{table*}[ht]
\centering
\begin{tabular}{@{}lcccccccccc@{}}
\toprule
Method & CRM & Dong & LIME & MF & Retinex-Net & NPE & GLAD & KinD++ & MIRNet-v2 & DARK \\
       & \cite{CRM2017ying} & \cite{Dong2011fast} & \cite{LIME2016guo} & \cite{MF2016weighted} & \cite{wei2018RetinexNet} & \cite{wang2013NPE} & \cite{wang2018GLAD} & \cite{zhang2021KIDpp} & \cite{Zamir2022} & (Ours) \\
\midrule
PSNR & 17.20 & 16.72 & 16.76 & 18.79 & 16.77 & 16.97 & 19.72 & 21.30 & \textbf{24.74} & \underline{21.16}\\
SSIM & 0.644 & 0.582 & 0.564 & 0.642 & 0.559 & 0.589 & 0.703 & 0.822 & \textbf{0.851} & \underline{0.767}\\
\bottomrule
\end{tabular}
\caption{Low-light image enhancement evaluation on the LoL dataset\cite{loldata}. The proposed method significantly advances the state-of-the-art.}
\label{tab:psnr}
\end{table*}

\subsection{Loss functions and Metrics}

In the proposed image restoration model, the loss function plays a crucial role in guiding the network towards generating high-quality reconstructions. The model utilizes a combination of weighted loss functions built in basicSR \cite{basicsr}, namely L1 (mean absolute error), MSE (mean squared error), PSNR (Peak Signal-to-Noise Ratio), and Charbonnier loss. Each of these loss functions is wrapped with a weighted\_loss decorator that allows the integration of element-wise weights, enhancing the flexibility to focus on specific areas or features within the images, such as edges or textures.

The L1 Loss is implemented to minimize the average absolute differences between the predicted and target images, providing robustness against outliers. 

The MSE Loss further refines the model by penalizing the squared discrepancies between the predicted outputs and the ground truths. This loss function emphasizes larger errors more significantly than smaller ones, which can be particularly useful for ensuring fidelity in regions with high-error values.

The PSNR Loss, specifically tailored for image processing, measures the peak error. The implementation modifies the standard PSNR approach by possibly converting images into a luminance-only format before computing the error, making it highly suitable for scenarios where human visual perception is prioritized in grayscale image contexts.

Lastly, the Charbonnier Loss provides a smooth approximation to L1 loss by incorporating a small constant. This loss is particularly effective in preserving image details and reducing artifacts in the restored images.

Together, these loss functions form a comprehensive loss landscape that not only penalizes the pixel-wise errors but also enhances the perceptual quality of the restored images, making the model adept at handling various image degradation patterns encountered in real-world scenarios.

In assessing the performance of the image restoration model, two widely recognized metrics are employed: Peak Signal-to-Noise Ratio (PSNR) and Structural Similarity Index Measure (SSIM). PSNR is instrumental in evaluating the fidelity of the restored images compared to the original, undegraded images by measuring the maximum error between them. It is particularly valuable in contexts where precise pixel-wise accuracy is critical. On the other hand, SSIM assesses the similarity in terms of luminance, contrast, and structure between the reconstructed and the original images, thereby offering insights into the perceptual quality of the restoration. Together, PSNR and SSIM provide a robust framework for quantitatively measuring both the technical and perceptual effectiveness of our low-light enhancement model, ensuring that it meets both objective and subjective quality standards.

\subsection{Hyperparameter Configuration}

Here we detail the hyperparameter settings employed across various components of our model.

\begin{itemize}
    \item \textbf{DataLoader}: The DataLoader was configured to shuffle data and utilize a batch size of 4. Progressive training was employed with iterations divided into stages with decreasing batch sizes from 8 to 1. Model progressively handled larger image patches, from $128\times128$ to $384\times384$ pixels.
    \item \textbf{Network}: The network comprises an input and output of 3 channels each, designed to process RGB images. It features 80 base features, scaled by a factor of 1.5.
    \item \textbf{Optimizer}: An Adam optimizer was employed with an initial learning rate of 2e-4, along with standard beta values of 0.9 and 0.999. The learning rate schedule was split into two cycles: a fixed rate for the initial 46,000 iterations, followed by a cosine annealing strategy decreasing the rate to 1e-6 over 100,000 iterations.
    \item \textbf{Augmentation}: To prevent overfitting and introduce regularization, the model was configured with mixup augmentation with a beta parameter of 1.2.
\end{itemize}

\section{Result Analysis}

\begin{figure}[ht]
    \centering
    % Row 1
    \begin{minipage}{0.33\linewidth}
        \includegraphics[width=\linewidth]{img/input_547.png}% First image in row 1
    \end{minipage}\hfill
    \begin{minipage}{0.33\linewidth}
        \includegraphics[width=\linewidth]{img/result_547.png} % Second image in row 1
    \end{minipage}\hfill
    \begin{minipage}{0.33\linewidth}
        \includegraphics[width=\linewidth]{img/true_547.png} % Third image in row 1
    \end{minipage}

    % Row 2
    \begin{minipage}{0.33\linewidth}
        \includegraphics[width=\linewidth]{img/input_748.png} % First image in row 2
    \end{minipage}\hfill
    \begin{minipage}{0.33\linewidth}
        \includegraphics[width=\linewidth]{img/result_748.png} % Second image in row 2
    \end{minipage}\hfill
    \begin{minipage}{0.33\linewidth}
        \includegraphics[width=\linewidth]{img/true_748.png} % Third image in row 2
    \end{minipage}

    % Row 3
    \begin{minipage}{0.33\linewidth}
        \includegraphics[width=\linewidth]{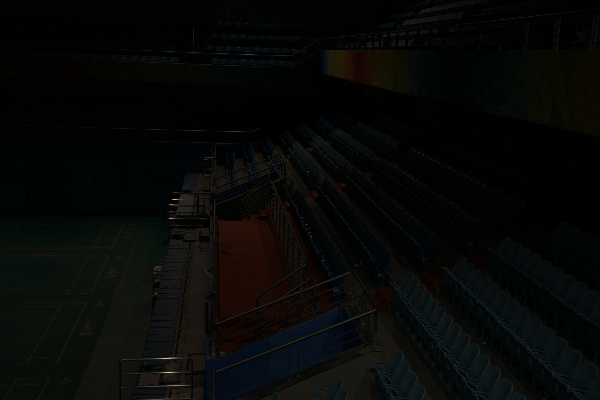} % First image in row 3
        \caption*{Input}
    \end{minipage}\hfill
    \begin{minipage}{0.33\linewidth}
        \includegraphics[width=\linewidth]{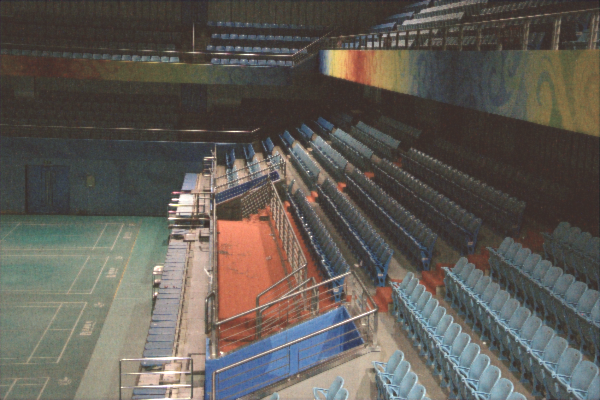} % Second image in row 3
        \caption*{Restored}
    \end{minipage}\hfill
    \begin{minipage}{0.33\linewidth}
        \includegraphics[width=\linewidth]{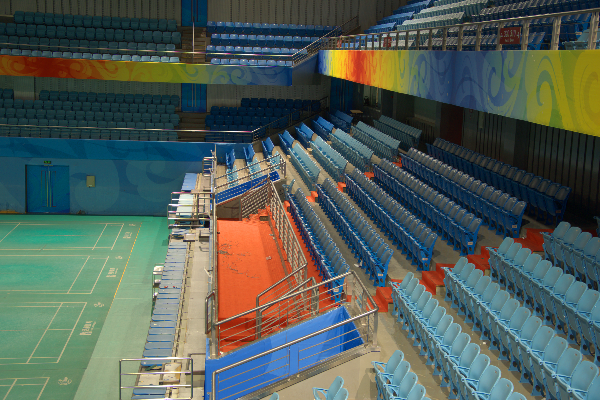} % Third image in row 3
        \caption*{Ground Truth}
    \end{minipage}
    \caption{Results on the validation set}
    \label{fig:results}
\end{figure}

\subsection{Validation}
Figure \ref{fig:results} visualizes our outputs on the LoL dataset\cite{loldata}.
The images show notable improvements and closely resemble the ground truth.
Missing details in the original images are effectively restored, featuring vibrant colors and appropriate illumination.

Our output images display diminished values across the RGB channels, resulting in a grayer appearance compared to the ground truth. This issue may stem from our decision not to utilize the illumination features generated by the estimator. Initial tests suggested that incorporating these illumination features did not markedly enhance the model's accuracy. Furthermore, we aimed to optimize our computational resources effectively.

\subsection{Comparison with Other Models}
\subsubsection{Model Size}

Our model is notably compact, containing a total of 187,123 parameters. This makes it significantly smaller for certain applications when compared to many other models in the field. MIRNet-v2 \cite{Zamir2022MIRNetv2} comprises approximately 5.9 million parameters while Model Retinexformer\cite{Cai_2023_ICCV} has 1.6 million parameters.

The vast difference in the number of parameters highlights our model's streamlined design, which is tailored to achieve efficient performance while maintaining a minimalistic architecture.

\subsubsection{Results}

This section highlights the performance of our algorithm by assessing it for image enhancement tasks.
We present the PSNR/SSIM scores of our approach alongside various other methods in Table \ref{tab:psnr} for the LoL dataset \cite{loldata}.
The results show that our model significantly outperforms earlier techniques.
Importantly, while our method, DARK, achieves performance comparable to the recent KinD++ \cite{zhang2021KIDpp} on the LoL dataset \cite{loldata}, it is considerably more efficient in terms of computational resources.

Nevertheless, in the context of larger models such as MIRNet-v2 \cite{Zamir2022MIRNetv2}, there remains room for improvement in our model's performance.

\subsection{Expectation}
We designed the model with the expectation that it would restore the overall features of the input image, accepting minor inaccuracies as permissible.
Our model exceeded our expectations by delivering a lightweight, low-light enhancement solution optimized for daily use.
It boasts small number of parameters, requires minimal training time, and is designed for laptop compatibility.
Moreover, the quality of the enhanced images is satisfactory, as evidenced by both the PSNR values and visual assessments conducted with the naked eye.

\section{Conclusion}
\subsection{Summary}
In this project, we conducted an extensive review of over ten research papers on low-light image enhancement to establish a robust knowledge base, with particular emphasis on MIRNet\_v2 \cite{Zamir2022MIRNetv2} and Retinexformer \cite{Cai_2023_ICCV} for further development. Inspired by the architectural design of MIRNet-v2 \cite{Zamir2022MIRNetv2} and employing the illumination estimation technique from Retinexformer \cite{Cai_2023_ICCV}, our model achieved a Peak Signal-to-Noise Ratio (PSNR) of 21.16 and a Structural Similarity Index (SSIM) of 0.767. Our light-weight model has 187,123 parameters in total. Training such a model with 100,000 iterations on the LoL datasets takes 80 minutes on an NVIDIA GeForce RTX 3060 Laptop GPU.

\subsection{Future plans}
There are several areas we aim to further investigate:
\begin{itemize}
\item \textbf{Advanced Denoiser:} Our current results still retain visible artifacts. 
As indicated by prior research such as Retinexformer \cite{Cai_2023_ICCV}, these artifacts largely originate from the inherently low illumination of the scene and are further exacerbated by the light-up process. We plan to develop a lightweight transformer-based denoiser module that utilizes the illumination features generated by the estimator. With additional time, we intend to further investigate this component to improve our model's effectiveness.

% According to previous study by Retinexformer \cite{Cai_2023_ICCV}, the artifacts mainly stem from the original low-illumination scene and corruptions amplified by the light-up process.
% We will implement a light-weight tranformer-based denoiser module using the illumination features generated by the estimator.
% Given more time, we would explore this component to enhance our model's capabilities.
% In Retinexformer, the illumination estimator outputs two layers: illu\_map (used to obtain lit-up image) and illu\_feat (used to restore the corruptions in the under-exposed scenes). We integrated illu\_map into our model for its ability to identify and adjust different parts of an image. Illu\_feat, which was used for denoising, was not implemented due to its large size. 
\item \textbf{Architecture Optimization:} Our innovative model has already shown satisfying results. However, there is potential to achieve even higher PSNR by optimizing the arrangement of blocks without increasing parameter count. Future efforts will focus on exploring different architectures and conducting comparative analyses.
\item \textbf{More Datasets:} Our training and testing were limited to the LoL dataset \cite{loldata}, which contains 485 training pairs and 15 testing pairs at a resolution of $400\times600$. We plan to test on additional datasets to better simulate various real-world scenarios and enhance the model's applicability for everyday use.
\end{itemize}

\subsection{Implications and Achievements}
This project demonstrates significant progress in low-light image enhancement. By applying the basicSR\cite{basicsr} framework, incorporating ideas from Retinexformer\cite{Cai_2023_ICCV} and MIRNet-v2\cite{Zamir2022MIRNetv2}, and designing our own architecture (MMRB) and customized blocks (SCB, SRCB), our model exceeds initial expectations, offering a practical solution for everyday use with efficient training and small parameter count. It achieves high-quality enhancements as confirmed by both technical metrics and visual assessments.

%%%%%%%%% REFERENCES
{\small
\bibliographystyle{ieee_fullname}
\bibliography{dark}
}

% \includepdf[pages=-]{appendix.pdf}

\end{document}